\documentclass[10pt, a4paper]{article}

\usepackage
{lrec-coling2024} 
\usepackage{xcolor}
\usepackage{graphicx}
\usepackage{caption}
\usepackage{subcaption}

\usepackage{booktabs}
\usepackage{multirow}
\usepackage{soul}
\usepackage{amsfonts}
\usepackage{amsmath,amssymb}
\usepackage{tikz}
\usetikzlibrary{shapes.misc}
\usepackage{float}
\usetikzlibrary{arrows.meta}
\usepackage{todonotes}
\usepackage{url}
\usepackage[T1]{fontenc}
\usepackage{amsfonts}
\usepackage{amssymb}
\usepackage[symbol]{footmisc}

\newtheorem{definition}{Definition}

\definecolor{PurpleColour}{HTML}{741b47}
\definecolor{PurpleColor2}{HTML}{8b0031}
\definecolor{OliveGreen}{HTML}{38761d}
\definecolor{BurntOrange}{HTML}{b45d07}

\newcommand{\Neema}[1]{\textcolor{magenta}{#1}}
\newcommand{\defif}{iff }
\newcommand{\freeX}{\ensuremath{\leadsto_f}}
\newcommand{\deductiveX}{\ensuremath{\leadsto_d}}
\newcommand{\argumentativeX}{\ensuremath{\leadsto_a}}

\title{Towards a Framework for Evaluating 
Explanations\newline in Automated Fact Verification}

\name{Neema Kotonya\textsuperscript{1,2}\footnotemark{*}\thanks{*This work is based on the author's doctoral thesis, completed at Imperial College London.} and Francesca Toni\textsuperscript{1}} 

\address{
          \textsuperscript{1}Department of Computing, Imperial College London;
         \textsuperscript{2}Dataminr, Inc.\\
         \{n.kotonya18,f.toni\}@imperial.ac.uk}

\abstract{
As deep neural models in NLP become more complex, and as a consequence opaque, the necessity to interpret them becomes greater. A burgeoning interest has emerged in rationalizing explanations to provide short and coherent justifications for predictions. In this position paper, we advocate for a formal framework for key concepts and properties about \emph{rationalizing} explanations to support their evaluation systematically. We also outline one such formal framework, tailored to rationalizing explanations of increasingly complex structures, from \emph{free-form} explanations to \emph{deductive} explanations, to \emph{argumentative} explanations (with the richest structure). Focusing on the \emph{automated fact verification}  task, we provide illustrations of the use and usefulness of our formalization for evaluating explanations, tailored to their varying structures.
\\ \newline  }

\begin{document}

\maketitleabstract

\section{Introduction}

In recent years, we have seen great performance success in natural language generation (NLG) and understanding (NLU), facilitated primarily by the use of sophisticated large language models (LLMs), e.g. LLaMA \citep{llama2023}. Despite these accomplishments, the complexity of these models calls for a greater need to interpret their computations. Interpretability of this kind would be desirable in numerous settings, e.g. some models are employed in safety and privacy critical applications~\cite{deza2021interpretability}, where it is important to understand whether these models are making the correct predictions for the right reasons \citep{mccoy-etal-2019-right}.

An increased focus on model interpretability has given way to several insightful works \cite{Belinkov-etal-2020-interpretability,madsen2021post}, exploring several angles including examining model robustness through the use of perturbations (e.g. adversarial attacks) \cite{song-etal-2021-universal}, and generating natural language explanations and evaluating their faithfulness \cite{jacovi-goldberg-2020-towards}.  The latter is the focus of this 
position paper.  Explanations extracted for deep neural models' predictions take a variety of forms. Earlier work employed explanations in the form of attention heat-maps and highlighted tokens  \cite{li-etal-2016-visualizing}. More recent work focuses on generating richer explanations, e.g. in graphical form \cite{thayaparan-etal-2021-explainable,saha-etal-2021-explagraphs,pmlr-v162-lampinen22a}, alongside techniques, similar to work in text generation and summarization, to 
obtain model faithful or label-consistent explanations \cite{kumar-talukdar-2020-nile,chrysostomou-aletras-2021-improving}.

\begin{figure}[ht]
    \centering
     \begin{subfigure}[b]{0.32\linewidth}
         \centering
         \includegraphics[width=\linewidth]{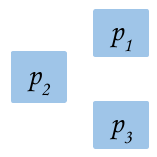}
         \caption{
        Free-form}
       
         \label{fig:unstructured-explanation}
     \end{subfigure}
     \hfill
     \begin{subfigure}[b]{0.32\linewidth}
         \centering
         \includegraphics[width=\linewidth]{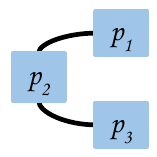}
         \caption{Deductive
         }
         \label{fig:structured-explanation}
     \end{subfigure}
     \hfill
     \begin{subfigure}[b]{0.32\linewidth}
         \centering
         \includegraphics[width=\linewidth]{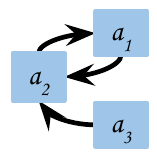}
         \caption{Argumentative}
         \label{fig:argumentative-explanation}
     \end{subfigure}
        \caption{Abstract illustrations of the three classes of explanations explored in this paper (where the p$_i$ are propositions and the a$_j$ are arguments).}
        \label{fig:three-formats-explanations-abstract}
\end{figure}

Here, we take the view that explanations are best understood as rationales for predictions, where these rationales are 
expressed in natural language, and typically extracted from the input text(s). The idea of rationalizing texts as explanations is discussed by \citet{lei-etal-2016-rationalizing} who deem rationales parts of the input text that are the most relevant for the predictions. This technique is similar to the processes humans undertake when delivering explanations to each other.  We take the view that, depending on context and human preferences, rationales should be in one of three different formats, as illustrated in Figure~\ref{fig:three-formats-explanations-abstract}.\footnote{Here, and in the remainder of this paper, illustrations are constructed by hand for space reasons, unless a source is explicitly given.}  These have already been proposed individually in some of the literature. For example \citet{atanasova-etal-2020-generating} construct free-form explanations in the form of summaries; deductive explanations are employed by \citet{krishna-etal-2022-proofver}; and contrastive explanations which can be modeled as argumentative frameworks are employed by \citet{derczynski-etal-2017-semeval,gorrell-etal-2019-semeval,schuster-etal-2021-get}. We systematize this line of work, by providing a paradigm in which we formally define three explanation formats, state their properties, and devise metrics for evaluating explanations of each format.

We focus on per-prediction \emph{rationalizing explanations}, which are specific to and offer reasons for individual predictions. 
We ground our analysis and paradigm on automated fact verification, for two main reasons. First, evidence-based automated fact verification is a knowledge-intensive task, thus it is important to provide explanations for predictions providing a summary of this evidence (and counter-evidence documents, if any). Second, the nature of misinformation claims is often polarizing, emotive, and hyper-partisan \cite{potthast-etal-2018-stylometric}, and thus 
explanations that are rationalizing can aid in gaining the end user's trust in the system delivering the predictions. Rationales may amount to free text (e.g. as in \citep{Camburu:2018}), or structured descriptions (e.g. as in \citep{tafjord-etal-2021-proofwriter}), possibly with an argumentative flavor (e.g. as in \citep{kotonya-toni-2019-gradual}, \citep{schuster-etal-2021-get}, \citep{chen-etal-2021-kace}, and \citep{dougrez-lewis-etal-2022-phemeplus}), present explanations as debates including competing (and possibly contrasting) arguments.

Concretely, we make the following contributions. 
\begin{itemize}
    \item 
First, we offer formal definitions of terms frequently employed in the literature on explanations for neural NLP. 
We present a paradigm for conceptualizing rationale-based explanations, expanding on \cite{wiegreffe-etal-2020-learning}, and viewing explanations by degree of structure (see Figure~\ref{fig:three-formats-explanations-abstract}).
\item Second, we argue the case for concrete metrics for evaluating such explanations based on properties. To this end, we propose and define 
several desirable properties for evaluating \emph{free-form}, \emph{deductive}, and \emph{argumentative} explanations. 
We then offer means 
for employing said properties for evaluation in empirical settings. 
\end{itemize}
Note that, while we restrict attention to (the evaluation of) explanations for predictions for the task of automated fact verification, the formalism, properties and metrics that we introduce can in principle be employed for evaluating explanations irrespective of the underlying task.

\section{Related Work}


The need for an explanation of NLP prediction tools is well established \cite{doshi2017towards,Ribeiro:2018,gohel2021explainable}
and rationalizations, as explanations are advocated by several \cite{rajani-etal-2019-explain,deyoung-etal-2020-eraser}.
In the context of automated fact verification, several bespoke forms of explanations have been proposed (see \cite{kotonya-toni-2020-explainable} for an overview), including rationalizations as explanations \cite{DBLP:conf/aaai/RanaKGSSR22,DBLP:conf/aaai/SiZZ23}.
Despite the many advances in rationalizing explainable NLP, we still observe the following shortcomings in the existing literature.

First, there are no agreed-upon definitions of what constitutes an explanation for an NLP prediction or the preferred methods for generating explanations. Many approaches are taken when characterizing and generating rationales for explanations. For example, \citet{deyoung-etal-2020-eraser} describe rationales as \textit{snippets that support the outputs} of a model. \citet{wadden-etal-2020-fact} consider rationales to be \textit{a minimal collection of sentences} the sum of which implies the veracity of a claim. \citet{schuster-etal-2021-get} extend this idea to include both supporting and contrastive evidence, i.e., rationales can favor one verdict or support an alternative verdict. \citet{ross-etal-2021-explaining} explore explanations as edits,  contrastive explanations in this case amount to the edits to the inputs which an alternative output. Contrastive explanations are closely related to counterfactual explanations \cite{guidotti2022counterfactual}.


Second, there is considerable work on identifying and evaluating properties of explanations in NLP \cite{jacovi-goldberg-2020-towards,atanasova-etal-2023-faithfulness}
but the focus of existing works is not on 
the evaluation of rationalizing explanations and prioritize properties can be interpreted as user requirements and are related to the relationship between the explanation and model prediction, e.g. faithfulness \cite{jacovi-goldberg-2020-towards,atanasova-etal-2023-faithfulness}, robustness \cite{10.1145/3447548.3470806} and sufficiency \cite{Chrysostomou_Aletras_2022}. Instead, the focus of our work is on properties related to explanation form, i.e. what new properties emerge as we enrich the structure of an explanation?
As there has been keen interest in the evaluation of deep NLP models \cite{ribeiro-etal-2020-beyond}, it would be valuable for this to extend to the evaluation of rationalizing explanations.

Third, there has been little effort to define and formalize a rigorous set of desirable criteria specific to rationalizing explanations. Some recent examples exist~\cite{Nauta2023}, but they do not focus on rationale-based explanations for fact-checking as we do and take a high-level approach to discuss properties, whereas we offer concrete definitions. 
An example of some effort in this direction for a free-form explanation for fact-checking is given in \citep{kotonya-toni-2020-explainable-automated}. \citet{atanasova-etal-2020-diagnostic} perform a diagnostic study of explainability for text classification concerning several properties. However, their focus is not specifically on rationalizing explanations; also they do not consider explainability in the automated fact verification context. This paper aims to outline a direction for addressing these issues.

\section{Definitions}

We define three classes of NLP (rationale-based) explanations, as abstractions of explanations found in the literature. The three classes amount to \emph{free-form} explanations (\S~\ref{ssec:unstructured-explanations}); \emph{deductive} explanations (\S~\ref{ssec:structured-explanation}), e.g. chains of facts as in \citep{yang-etal-2018-hotpotqa}; and \emph{argumentative} explanations (\S~\ref{ssec:argumentative-explanations}), e.g. providing reasons for supporting or refuting a claim as in \citep{wadden-etal-2020-fact} and \citep{schuster-etal-2021-get}. We assume that the end users of our proposed framework are humans, and this motivates us to consider explanation formats and properties that align closely with human explanations, following established views in Explainable AI (XAI), most notably by \citet{Miller19_survey,DBLP:journals/ker/Miller21}.

\subsection{Free-form Explanations}
\label{ssec:unstructured-explanations}
Free-form explanations are the most common explanatory outputs in the 
rationale-based landscape and reflect a large part of earlier literature on 
explaining deep neural models \cite{wiegreffe-etal-2020-learning}. We define them abstractly below, using a generic notion of \emph{proposition}. 
Note that, 
since we want to provide as general guidance as possible to designers of explanations, we do not place any stipulations on the nature of propositions but, in practice, the decision of what to admit as propositions needs to be taken before free-form explanations are drawn from models.
Concretely, propositions could amount, for example, to words or phrases (e.g. occurring in the input text being explained), model predictions (e.g. that the model predicts that the input is true), and tokens understood by the underlying models.

\begin{definition}
\label{def:free-informal}
A \textbf{free-form explanation}  amounts to a finite, non-empty sequence of propositions. A \textbf{free-form explanation for a model's prediction, given an input}, is a free-form explanation that includes, among its propositions, some elements of the input and the prediction itself. We use the notation $\mathcal{P} \freeX [m(\mathbf{X}) = \hat{y} ]$ to indicate that $\mathcal{P}$ is a free-form explanation for prediction $\hat{y}$ by model $m$, given input $\mathbf{X}$.
\end{definition}

The inclusion of the prediction in a free-form explanation relates to 
``relevance'' thereof to the prediction. This prediction is often implicit, as in the example presented in Table \ref{tab:example_free_form}.

\begin{table}[ht]
\centering
    \begin{tikzpicture}
        \node[inner sep=0pt, draw, rounded corners=.5em, fill=blue!30!cyan!10] (table) {
            \begin{tabular}{p{7cm}}
  \addlinespace[1ex]               
 A popular Facebook post about the life and death of British mathematician Alan Turing is truthful.\\
 \hline  \addlinespace[1ex]
 \textbf{Verdict: }  \texttt{Mostly True} \\
 \textbf{Explanation:} \\
   The popular Facebook post got most of the facts right $(p_1)$. However, there’s no evidence that Turing inspired the design of the Apple computer company’s logo $(p_2)$. Also, Turing’s death in 1954 deserves further examination than what was provided in the post, which we included below $(p_3)$.  \\  \addlinespace[1ex]
            \end{tabular} 
        };
        \draw[rounded corners=.5em] (table.south west) rectangle (table.north east);
    \end{tikzpicture}

\caption{Example of free-form explanation, matching the abstract illustration in Figure \ref{fig:three-formats-explanations-abstract}(\subref{fig:unstructured-explanation}). Here, the claim,  prediction (Verdict (label)) and explanation (except for the $p_i$, which are our addition) are taken from the claim verification platform Snopes 
\url{https://www.snopes.com/fact-check/alan-turing-facebook-post/}.
}
\label{tab:example_free_form}
\end{table}

Note that we do not enforce that $\mathcal{P}$ is restricted to elements in the input $\mathbf{X}$ and indeed, in general, it could also include elements not in $\mathbf{X}$, as in the illustration in Table~\ref{tab:example_free_form}. Free-form explanations may take several concrete forms in practice. \citet{wiegreffe-etal-2020-learning} distinguish between highlights, e.g. token-wise saliency maps,  and free-form text.
Neural attention is used to create saliency maps \cite{li-etal-2016-visualizing}. Instead,  
when free-form natural language explanations are used, a sequence-to-sequence model is typically employed to generate a text that serves as the rationale for the predictions. 
In this setting, 
\citet{Camburu:2018} explore two paradigms: one which jointly generates 
explanation and prediction, and another which first generates the explanation and then the prediction. As a further example, \citet{kumar-talukdar-2020-nile} looks to generate label-specific explanations for each possible label prediction. Furthermore,
\citet{kotonya-toni-2020-explainable-automated}'s explanations as summaries could be seen as a form of free-form explanation. 
Most explanations of this type are self-generated, i.e. the model is expected to both predict outputs \textit{and} explain its reasoning, e.g. by way of a prompt-based model \cite{Narang2020WT5TT,marasovic2021few}. 

However, there is an ongoing debate regarding whether explanations of this form are sufficient \cite{jain-wallace-2019-attention,wiegreffe-pinter-2019-attention,pruthi-etal-2020-learning}, 
to which we contribute by characterizing other forms, in \S~\ref{ssec:structured-explanation}-\ref{ssec:argumentative-explanations} below.

\subsection{Deductive Explanations}
\label{ssec:structured-explanation}

Like free-form explanations, deductive explanations also consist of propositions, but these propositions are ``connected'' by a (binary) \emph{relation}. Again, since we aim to provide as general guidance as possible to designers of explanations, we do not place any stipulations on the nature of $\mathcal{R}$ and keep it abstract, but, in practice, the decision of what to admit as relation needs to be taken before deductive explanations are drawn from models. 
Concretely, for example, this relation could be support between propositions, or relevance between propositions, or a proposition being in relation with another could amount to the former being a reason for the latter or for the two to have something in common. Deductive explanations also have strong parallels with the reasoning obtained through chain-of-thought \cite{wei2022}, tree-of-thought \cite{Yao2023}, and other similar graph topological large language model prompting techniques.

\begin{definition}
\label{def:deductive}
A \textbf{deductive explanation} amounts to
 a pair composed of a finite, non-empty sequence of propositions and a binary relation over the propositions. A \textbf{deductive explanation for a model prediction, given an input}, is a deductive explanation that includes, among the pair described above
 its components, some elements of the input as well as the prediction itself. We use the notation $\langle \mathcal{P},\mathcal{R}\rangle \deductiveX [m(\mathbf{X}) = \hat{y} ]$ to indicate that $\langle \mathcal{P},\mathcal{R}\rangle$ is a deductive explanation for prediction $\hat{y}$ by model $m$, given input $\mathbf{X}$.
\end{definition}

\begin{table}[H]
\centering
    \begin{tikzpicture}
        \node[inner sep=0pt, draw, rounded corners=.5em, fill=blue!30!cyan!10]  (table) {
            \begin{tabular}{p{7cm}}
             \addlinespace[1ex]
$p_3$: A daffodil plant can live for more than two years.\\
\hline 
 \addlinespace[1ex]
     \textbf{Verdict:} \texttt{Verified}\\
     \textbf{Explanation $\langle \mathcal{P, R} \rangle$}, where: \\
      $\mathcal{P} = \{p_1, p_2, p_3\}$, for:\\
            $p_1$: Daffodil is the common name for plants of the narcissus genus, which are perennial.    \\
            $p_2$: 
      A perennial plant has a minimum life span of two years.   \\

    \textit{
    $\mathcal{R} = \{(p_1, p_2), (
    p_2,p_3)\}.$}\\
     \addlinespace[1ex]
            \end{tabular}
        };
        \draw[rounded corners=.5em] (table.south west) rectangle (table.north east);
    \end{tikzpicture}
    \caption{Example of deductive explanation, matching the abstract illustration in Figure~\ref{fig:three-formats-explanations-abstract}(\subref{fig:structured-explanation}).}
    \label{tab:deductive_example}
\end{table}

The example in Table \ref{tab:deductive_example} gives an illustration of deductive explanation: here and later, we represent $\mathcal{R}$ as a set of pairs, so, for example, $(p_1,p_2) \in \mathcal{R}$ indicates that $p_1$ and $p_2$ are related by $\mathcal{R}$. 
Here, $\mathcal{R}$ may be seen as a logical reasoning chain or as a linking of propositions by common entities. 
In general, several other possibilities could be considered for identifying $\mathcal{R}$, e.g. chronological ordering of evidence.

As in the case of free-form explanations, we enforce ``relevance'', ensuring the prediction is in $\mathcal{P}$ (but, again, we may have that the prediction is implicit, as in Table~\ref{tab:deductive_example}). 

Also, we may impose a direction in $\mathcal{R}$ or not, if we want to capture bidirectionality as in the case of $\mathcal{R}$ representing that propositions have something in common.  
Furthermore, a compound layered deductive explanation could be acquired by considering multiple semantics for $\mathcal{R}$: we leave this as future work.

A clear form of deductive explanations in the literature is \textit{chains of connected facts} \cite{inoue-etal-2020-r4c,tafjord-etal-2021-proofwriter}. The example we present in Table \ref{tab:deductive_example} amounts to a chain of facts because there
is a sequence of propositions $p_1 \rightarrow p_2 \rightarrow p_3$ that leads from the proposition $p_1$ to the claim ($p_3$). Propositions $p_1$ and $p_2$ provide evidence for which the logical conclusion is $p_3$, thus this chain of facts justifies the claim $p_3$. Chains of facts are analogous to chain-of-thought prompting \cite{DBLP:conf/nips/Wei0SBIXCLZ22,lampinen-etal-2022-language}. In our view, the reasoning output produced by the chain-of-thought process amounts to a deductive explanation.

\subsection{Argumentative Explanations}
\label{ssec:argumentative-explanations}
We now examine explanations that provide justifications for model predictions using \emph{arguments}, as opposed to simple propositions.
Intuitively, an argument consists of a conclusion that is supported by premises.  In particular, we can choose premises and conclusions of arguments to be propositions, understood broadly as in the definitions of free-form and deductive explanations. 

Since the kinds of arguments used in argumentative explanations are expressed in natural language, we do not place any stipulations on the logical connection between premises and the conclusion of an argument.  
In particular, arguments could be enthymemes \cite{razuvayevskaya2017finding} with partially specified or even empty premises.

Arguments are the building blocks of debates. For this work, in the spirit of \cite{Dung:1995,AImag}, we represent debates as argumentation frameworks, modeling the interactions between arguments as relations.
Specifically, we focus on bipolar argumentation frameworks \cite{cayrol2005acceptability}, where an argument can be 
attacked or supported by arguments, thus modeling both conflict and agreement (respectively) between arguments. 
In the spirit of \cite{cayrol2005acceptability}, we leave the definition of what conflict or agreement may mean completely unspecified, assuming instead that they are captured by abstract relations.
We will return to them later in Definition~\ref{def:attsupp}. 

\begin{definition}
\label{def:argumentative-explanation}
An \textbf{argumentative explanation} is given by a 3-tuple which amounts to a finite, non-empty set of arguments, a binary attack relation over the set of arguments and a binary support relation over the set of arguments. An \textbf{argumentative explanation for
a model prediction, given an input}, is an argumentative explanation that includes, among its arguments, attack and support relations, some elements of the input as
well as the prediction itself. We use the notation
$\langle \mathcal{A},\mathcal{R_{\text{Sup}}}, 
\mathcal{R_{\text{Att}}} \rangle \argumentativeX [m(\mathbf{X}) = \hat{y}]$ to indicate that $\langle \mathcal{A},\mathcal{R_{\text{Sup}}}, 
\mathcal{R_{\text{Att}}} \rangle$ is a
an argumentative explanation for prediction $\hat{y}$ by model $m$, given input $\mathbf{X}$. 
\end{definition}

Definition \ref{def:argumentative-explanation} 
could be realized so that
at least one argument exists that admits as its conclusion the output prediction $\hat{y}$ and that some arguments need to admit propositions in the input among their premises. If there is just one argument in the argumentative explanation then these restrictions impose that the argument `connects' the input and the prediction, i.e. there is a way to reason from the inputs which forms a rationale for the prediction.

Definition~\ref{def:argumentative-explanation} leaves the attack/support relations unspecified. These could be defined in several ways, including as follows, making use of generic notions of \emph{contradiction} and \emph{implication}, 
as in \emph{textual entailment} \cite{dagan_dolan_magnini_roth_2010}.

\begin{definition}
Let $a_i, a_j \in \mathcal{A}$ where $\mathcal{A}$ is a set of arguments. 
Then: 
\begin{itemize}
\item $a_j$ \textbf{attacks} $a_i$ (\textbf{by undercutting}) if 
the conclusion of $a_j$ is in contradiction with some 
premise(s) 
in $a_i$; 
    \item 
$a_j$ 
\textbf{attacks} $a_i$ (\textbf{by rebutting}) if the conclusion of $a_j$ is in contradiction with the conclusion of ${a_i}$;  
\item 
$a_j$ \textbf{supports}  $a_i$ (\textbf{by providing reasons}) if the conclusion of $a_j$ implies some premise(s) of $a_i$;
\item $a_j$ \textbf{supports} $a_i$ (\textbf{by accrual}) if the conclusion of $ {a_j}$ implies  the conclusion of ${a_i}$.
\end{itemize}
\label{def:attsupp}
\end{definition}

Note that this definition borrows some concepts from the literature on argumentation frameworks, specifically, the notion of accrual \cite{prakken2005study}, undercutting \cite{bex2003towards}, and rebuttal \cite{kowalski1996abstract}.
%
Note also that typically attack by rebutting and support by accrual will be symmetric (
i.e. if $a_j$  attacks by rebuttal or supports by accrual  $a_i$, then $a_i$ does 
so towards $a_j$). For an illustration of argumentative explanations and notions of attack/support, consider Table \ref{tab:example_argumentative}. 

\begin{table}[ht]
\centering
    \begin{tikzpicture}
     \node[inner sep=0pt, draw, rounded corners=.5em, fill=blue!30!cyan!10] (table) {
    \begin{tabular}{p{7cm}}
     \addlinespace[1ex]
      \textit{
      Consider the following claim for fact-checking:}\\
 \addlinespace[1ex]
The King of the United States of America lives in the White House.\\\\

\textit{Also, consider the following arguments in the context of some argumentative explanation for prediction \texttt{Refuted} in verdict to the claim:} \\\\

    $a_1$: 
    the King of the USA does not live in the White House ($\hat{y}$)
because the USA has no king, as it is a republic ($p_1$).\\
    $a_2$: The White House is the official residence of the 
    USA
    President 
    ($p_2$), 
    thus it can not be the official residence of a king or any other head of state ($\hat{y}$).\\
    $a_3$: The head of state of the USA is the President ($c_3$
    ) because the title ``president'' is typically given to the head of a republic ($p_3$).\\\\

    These arguments can also be represented symbolically 
as pairs (consisting of premises and claims): $a_1=(\{ p_1\},\hat{y}), a_2=(\{p_2\}, \hat{y}), a_3=(\{p_3\}, c_3)$.\\
 \addlinespace[1ex]
    \end{tabular}
            };
    \draw[rounded corners=.5em] (table.south west) rectangle (table.north east);
    \end{tikzpicture}
    \caption{Concrete example of arguments in an argumentative explanation.}
    \label{tab:example_argumentative}

    \end{table}
    
Here, arguments $\mathcal{A} = \{a_1, a_2, a_3  \}$, and several argumentative explanations $\langle\mathcal{A}, \mathcal{R}_{\text{Sup}}, \mathcal{R}_{\text{Att}}\rangle$ 
are possible, including where:
\begin{enumerate}
    \item 
$\mathcal{R}_{\text{Sup}}= \{(a_3,a_2)\}$ and $\mathcal{R}_{\text{Att}} = \emptyset$; 
here,  argument $a_3$ is reinforcing argument $a_2$, we have adopted a view of support as providing a rationale to justify the conclusion of an argument; there are no attacking arguments in this particular explanation;
\item 
$\mathcal{R}_{\text{Sup}} = \{(a_3,a_2), (a_1,a_2), (a_2,a_1)\}$ and (again) $\mathcal{R}_{\text{Att}} = \emptyset$; here, 
arguments $a_1$ and $a_2$ corroborate one another (by accrual) in support of the output of the classifier. An abstract depiction of this argumentative explanation is shown in Figure \ref{fig:three-formats-explanations-abstract}(\subref{fig:argumentative-explanation}), with edges representing support.
\end{enumerate}

Note that in this example the explanation includes an argument ($a_3$) which is neither for the prediction nor for any alternative predictions. In some settings, argumentative explanations could be restricted to make sure that the conclusion of each argument in $\mathcal{A}$ must imply a prediction from the set of possible outputs for the model. 
For these types of argumentative explanations, which we may call \textit{flat} if there are arguments for different outcomes than computed by the classifier, the attack relation would be non-empty and include some attacks by a rebuttal. This may be the case, for example, in \citep{wadden-etal-2020-fact,schuster-etal-2021-get}. Whereas it is clear why we may want to include support, some considerations about the inclusion of attack in argumentative explanations are in order. We include an attack to reflect two scenarios:

\begin{enumerate}
    \item First, the possibility of conflicting evidence the model found in the input, giving reasons for undercutting or rebutting other 
    reasons;
    \item Second, 
    to represent the fact/foil relationship \cite{barnes1994p}, 
    given that attacks can distinguish between inputs which contribute to the 
    prediction from distractor inputs that do not; 
    here, attacks are of the rebuttal variety (as they point to contradictory predictions).
\end{enumerate}

In other words, attacks may be needed to explain a low-confidence prediction from a model.


\section{
Properties}
\label{sec:properties}
We define a number of properties for our forms of rationalizing explanations. Our list of properties is not exhaustive, and we see each of the properties as a useful criterion for assessing the validity of explanations in rationalizing a model's prediction in NLP. We consider separately properties tailored to free-form, structured and argumentative explanations (\S~\ref{ssec:properties-unstructured}, \ref{ssec:properties-structured} and \ref{ssec:properties-argumentative}, respectively).

\subsection{Free-Form Properties}
\label{ssec:properties-unstructured}
We adapt the properties for free-form explanations introduced by \citet{kotonya-toni-2020-explainable-automated} for evaluating explainable summaries (a form of free-form explanation) 
for automated fact verification. Here, we propose a single property for any free-form explanations:  \emph{coherence}. We define the notion of coherence in terms of a notion of logical contradiction (which could amount to the implication of negation), in line with the definitions in \citep{kotonya-toni-2020-explainable-automated}.

\begin{definition}
 A free-form explanation $\mathcal{P}$ satisfies \textbf{coherence} if there exists no 
 contradictory subset of propositions in $\mathcal{P}$. 
\end{definition}

Thus, coherence is a measure of the cohesiveness of propositions in a free-form explanation. For 
coherence to hold, in particular, any two propositions in an explanation must not contradict one another, i.e. there is no pairwise disagreement between propositions which make up the explanation. More generally, our definition excludes contradictions involving any number of propositions making up the explanation.


\subsection{Deductive Properties}
\label{sec:properties-structured}
\label{ssec:properties-structured}

The coherence property  for free-form explanations is 
still applicable to deductive explanations $\langle \mathcal{P}, \mathcal{R} \rangle$ on the set of propositions $\mathcal{P}$.
In addition, we identify four bespoke properties for deductive explanations: \emph{non-circularity}, \emph{(weak} and \emph{strong) relevance}, and \emph{non-redundancy}, defined below.


\begin{definition}
A deductive explanation $\langle\mathcal{P},\mathcal{R}\rangle$ 
is \textbf{non-circular} if there does not exist a proposition $p_i$ in 
$\mathcal{P}$ and a set of propositions $\mathcal{P}'$ from $\mathcal{P}$,
$\mathcal{P}'=\{p_1', \dots, p'_k\} \!\subseteq \! \mathcal{P}$, such that
$\{(p_1',p_2'),
\dots, (p_{k-1}',p_k')\} \!\subseteq \! \mathcal{R}$  and $p_1'=p_k'=p_i$.


\end{definition}

Thus,  non-circularity amounts to acyclicity of the $\mathcal{R}$ component of deductive explanations (when seeing them as directed graphs with propositions as nodes and elements of the relation as edges, as in Figure~\ref{fig:three-formats-explanations-abstract}). 
It seeks to avoid circular explanations, which are not sound in a rhetorical sense. 


\begin{definition} 
A deductive explanation  $\langle\mathcal{P},\mathcal{R}\rangle$ such that $\langle \mathcal{P},\mathcal{R}\rangle \deductiveX [m(\mathbf{X}) = \hat{y} ]$
is  \textbf{strongly relevant} if all propositions $p_i$ in $\mathcal{P}$  are such that $(p_i, \hat{y})$ is in $\mathcal{R}$.
\end{definition}

Thus, all propositions in a strongly relevant deductive explanation are directly connected to the model's prediction. 

\begin{definition}
A deductive explanation 
$\langle\mathcal{P},\mathcal{R}\rangle$ such that $\langle \mathcal{P},\mathcal{R}\rangle \!\deductiveX \![m(\mathbf{X}) \!=\! \hat{y} ]$
is \textbf{weakly relevant} 
if, for all propositions $p_i$ in $\mathcal{P}$,
there exists a set of propositions $\mathcal{P}'$ in $\mathcal{P}$, 
$\mathcal{P}'\!=\!
\{p_1', \dots, p'_k\} \!\subseteq\! \mathcal{P}$, 
such that
$\{(p_1',p_2'),
\dots, (p_{k-1}',p_k')\} \subseteq  \mathcal{R}$,
$p_1'=p_i$ and
$p_k'=\hat{y}$.
\end{definition}

Namely, for a deductive explanation to be weakly relevant, each proposition needs to be connected by some chain (path) to the prediction. Thus, in weakly relevant deductive explanations there are no unconnected propositions.
Note that we could easily define additional versions of relevance, e.g. to enforce links to input propositions 
(we refrain from doing so for lack of space). 

Non-redundancy, the last property we define for deductive explanations, requires that no superfluous propositions are contained in an explanation. 

\begin{definition}
A deductive explanation 
$\langle\mathcal{P},\mathcal{R}\rangle$ such that $\langle \mathcal{P},\mathcal{R}\rangle \deductiveX [m(\mathbf{X}) = \hat{y} ]$
is \textbf{non-redundant} \defif for all propositions $
p_i \in \mathcal{P} \setminus \{\hat{y}\}
$, for $\mathcal{P}' = \mathcal{P} \setminus \{p_i\}$ and $\mathcal{R}'=\mathcal{R}\cap (\mathcal{P}' \times \mathcal{P})$, the pair $\langle \mathcal{P}', \mathcal{R}'\rangle$ is not a deductive explanation for $\hat{y}$ by $m$, given $\mathbf{X}$,  i.e. $\langle \mathcal{P}', \mathcal{R}'\rangle \not\models m(\mathbf{X}) \rightarrow \hat{y}$.
\end{definition}

In other words, no proposition can be eliminated from a non-redundant explanation while still rationalizing the prediction for which it is intended.
Note that, when eliminating a proposition from a deductive explanation, we also delete from the relation component all connections to and from deleted propositions. 
Note also that, in practice, by our definition of deductive explanation,
$ \langle \mathcal{P}', \mathcal{R}'\rangle \not\models m(\mathbf{X}) \rightarrow \hat{y}$ means that either 
$\hat{y}$ is not in $\mathcal{P}'$ or $\mathcal{P}'$ contains no elements of the input $\mathbf{X}$.







\subsection{Argumentative Properties}

\label{sec:properties-argumentative-explanation}
\label{ssec:properties-argumentative}

The coherence property for free-form explanations, presented in \S~\ref{ssec:properties-unstructured}, could be enforced on the premises of arguments in argumentative explanations (e.g. by seeing arguments with premises $\mathcal{P}$ 
as free-form explanations $\mathcal{P}$). Furthermore, we could enforce properties similar in spirit to those for deductive explanations, presented in \S~\ref{sec:properties-structured}, to individual arguments in argumentative explanations, 
(e.g. by seeing arguments with premises $\mathcal{P}$ and conclusion $c$ as
  deductive explanation $\langle \mathcal{P}, \{(p,c) \mid p \in \mathcal{P}\} \rangle$).    
Here, we focus instead on additional properties of argumentative explanations.
Specifically, we identify a number of properties regarding relations between arguments.

First, we can impose 
natural conditions on the attack and support relations, i.e. that there are no cycles therein:

\begin{definition}
An argumentative explanation
$\langle \mathcal{A},\mathcal{R_{\text{Sup}}},\mathcal{R_{\text{Att}}} \rangle$
is \textbf{dialectically non-circular} if there is no argument $a_i$ in  $\mathcal{A}$ and no set of arguments $\mathcal{A}'$ from $\mathcal{A}$, 
$\mathcal{A}'=\{a_1', \dots, a'_k\}\subseteq \mathcal{A}$,  such that $a_1'=a_k'=a_i$, and 
$\{(a_1',a_2'), \dots, (a_{k-1}',a_k')\} \!\subseteq \! (\mathcal{R}_\text{{Sup}} \cup \mathcal{R}_{\text{Att}})$.
\end{definition}

\begin{figure}
    \centering
    \begin{subfigure}[b]{0.23\textwidth}
        \centering
        \includegraphics[width=\textwidth]{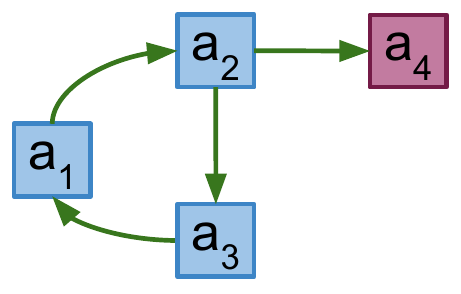}
        \caption{All supports.}
        \label{fig:subfigA}
    \end{subfigure}
    \hfill
    \begin{subfigure}[b]{0.23\textwidth}
        \centering
        \includegraphics[width=\textwidth]{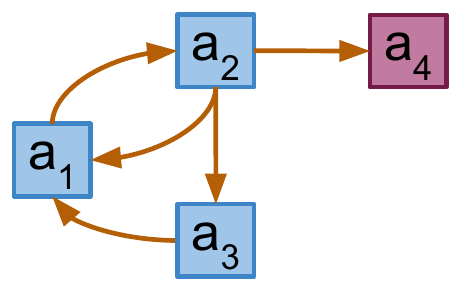}
        \caption{All attacks.}
        \label{fig:all-attacks}
    \end{subfigure}
    \hfill
        \begin{subfigure}[b]{0.23\textwidth}
        \centering
        \includegraphics[width=\textwidth]{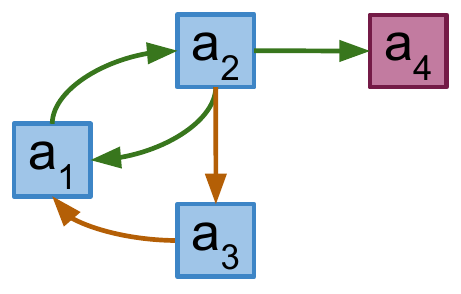}
        \caption{Mixed supports\newline and attacks.}
        \label{fig:mixed-attacks-supports}
    \end{subfigure}
    \caption{Examples of dialectical circularity for three argumentative explanations.  Attacks are shown in {\color{BurntOrange} orange} and supports are shown in {\color{OliveGreen} green}. Argument $a_4$ with conclusion $\hat{y}$ is {\color{PurpleColour}purple}.}
    \label{fig:dialectical_circularity}
\end{figure}

Violation of dialectical non-circularity would entail a cycle of supports and/or attacks.
The specific case of violation of dialectical non-circularity, when some argument in the explanation is self-supporting, may amount to a situation of an argument that is not grounded in evidence.
The specific case of violation of dialectical non-circularity by a cycle of supports (Figure~\ref{fig:dialectical_circularity}(a))  may amount to an 
unsound debate in a rhetorical sense.
The specific case of violation of dialectical non-circularity 
when some argument in the explanation is 
self-attacking 
may amount to a paradoxical situation of the argument 
being self-contradictory. 
 The specific case of 
 violation of dialectical non-circularity by
 a cycle of attacks (Figure~\ref{fig:dialectical_circularity}(b)) may also amount to an 
unsound debate in a rhetorical sense.
Mixed cases, as in Figure~\ref{fig:dialectical_circularity}(\subref{fig:mixed-attacks-supports}) are also challenging from a rhetorical perspective.

In addition, we can demand that argumentative explanations satisfy other dialectical properties, in the spirit of
various argumentation frameworks from symbolic AI, notably abstract argumentation \cite{Dung:1995}, bipolar argumentation \cite{cayrol2005acceptability} 
and quantified bipolar argumentation \cite{Baroni2019FromFP}.
These frameworks 
rely upon notions of
\emph{acceptability} of sets of arguments \cite{Dung:1995} or 
\emph{dialectical strength} for arguments \cite{Baroni2019FromFP}. In our setting, properties inspired by these notions can be used to point towards the explanations' credibility in the context of the confidence of the underlying model in the prediction. 
Intuitively, 
a credible (or strong) argument is supported by other (credible) argument(s). Conversely, a less credible (weakened) argument is attacked by (credible) arguments.
In this spirit, we deem an argumentative explanation \emph{dialectically faithful} if its credibility reflects the prediction confidence. We formalize this property using a generic notion of dialectical strength for arguments \cite{Baroni2019FromFP}: 

\begin{definition}
An argumentative explanation $\langle \mathcal{A},\mathcal{R_{\text{Sup}}},\mathcal{R_{\text{Att}}} \rangle$ such that  $\langle \mathcal{A},\mathcal{R_{\text{Sup}}}, \mathcal{R_{\text{Att}}} \rangle \argumentativeX [m(X) = \hat{y}]$ is \textbf{dialectically faithful} if

\begin{itemize}
\item whenever $m$ gives $\hat{y}$ with top confidence,
$\mathcal{R_{\text{Att}}}$ is such that there are no arguments in 
$\mathcal{A}$
attacking   by rebutting any argument for $\hat{y}$;
\item whenever $m$ gives $\hat{y}$ with high confidence, the dialectical strength of the arguments in $\mathcal{A}$ with conclusion $\hat{y}$ is higher than  the dialectical strength of the arguments attacking them (as per $\mathcal{R_{\text{Att}}}$);
\item whenever $m$ gives $\hat{y}$ with low confidence,  $\mathcal{A}$ must either include only dialectically weak arguments  with conclusion $\hat{y}$
or include some arguments 
attacking by rebutting some argument  for $\hat{y}$  %
with higher  dialectical strength than arguments in $\mathcal{A}$ with conclusion $\hat{y}$, if any.
\end{itemize}
\end{definition}

\begin{figure*}[ht]
    \centering
    \includegraphics[width=0.65\linewidth]{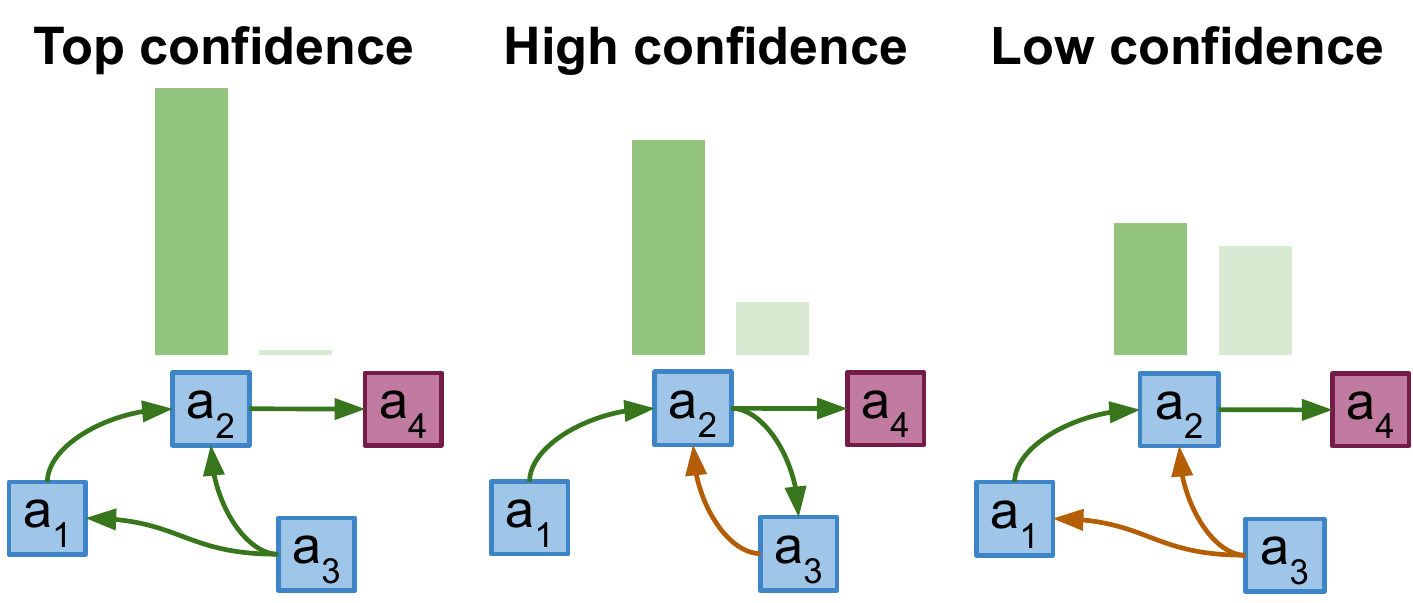}
    \caption{An illustration of 
    argumentative explanations for top, high, and low confidence (binary) predictions. 
    Attacks are shown in {\color{BurntOrange} orange} and supports are shown in {\color{OliveGreen} green}. Argument $a_4$ with conclusion $\hat{y}$ is {\color{PurpleColour}purple}.}
    \label{fig:dialectical-faithfulness-example}
\end{figure*}

Intuitively, the argument for a prediction with high confidence should be supported by 
strong arguments (as in the case of the argumentative explanations outlined 
when discussing Table \ref{tab:example_argumentative}).
Furthermore, consider the argumentative explanations in Figure \ref{fig:dialectical-faithfulness-example}. 
Here, 
the argument $a_4$ for the prediction with top confidence (left-most argumentative explanation) is not attacked, thus the explanation is dialectically faithful. 
Note that rebuttals imply contradiction between arguments' conclusions and are thus singled out in our definition of dialectical faithfulness. 
Also, in Figure \ref{fig:dialectical-faithfulness-example}, the argument $a_4$ for the prediction with high confidence (middle argumentative explanation), being again unattacked, trivially has a higher dialectical strength than its attackers, so, again, the explanation is dialectically faithful. 
Finally, 
the argument $a_4$ for the prediction with low confidence (right-most argumentative explanation) is supported by  argument $a_2$
which is weakened by the attack from $a_3$; the latter also weakens the support from $a_1$ to $a_2$; thus, overall $a_2$ is a weak argument and the argumentative explanation can be deemed dialectically faithful.

In the special case when dialectical strength is ``binary'' (in that it sanctions an argument as winning or losing, e.g. as in \citep{Dung:1995}),
we can refine dialectical faithfulness to define a notion of \emph{acceptability} as follows:

\begin{definition}
An argumentative explanation $\langle \mathcal{A},\mathcal{R_{\text{Sup}}},\mathcal{R_{\text{Att}}} \rangle$ such that  $\langle \mathcal{A},\mathcal{R_{\text{Sup}}}, \mathcal{R_{\text{Att}}} \rangle \argumentativeX [m(\mathbf{X}) = \hat{y}]$ is \textbf{acceptable} if

\begin{itemize}
\item 
whenever $m$ gives $\hat{y}$ with top or high confidence, there exists no $\mathcal{A'} \subseteq \mathcal{A}$ containing  
all arguments in $\mathcal{A}$ with conclusion $\hat{y}$
such that
 for all arguments $
 a_i \in \mathcal{A}'$, if  $a_j \in \mathcal{A}$ attacks $a_i$ (i.e. $(a_j,a_i) \in \mathcal{R}_{Att}$), then there exists $a_k \in \mathcal{A}'$ attacking $a_j$ (i.e. $(a_k,a_j) \in \mathcal{R}_{Att}$); in simpler terms, $\mathcal{A'}$ defends itself against all attacking arguments;

\item whenever $m$ gives $\hat{y}$ with bottom or low confidence,
for 
$ \mathcal{A'} \subseteq \mathcal{A}$ containing  
all arguments in $\mathcal{A}$ with conclusion $\hat{y}$, there exists
some arguments $ a_i \in \mathcal{A}'$ and $a_j \in \mathcal{A} \setminus \mathcal{A}' $ such that $a_j$ attacks $a_i$ (i.e. $(a_j,a_i) \in \mathcal{R}_{Att}$) but there exists no argument $a_k \in \mathcal{A'}$ such that $a_k$ attacks $a_j$ (i.e. $(a_k,a_j) \in \mathcal{R}_{Att}$);  in simpler terms,  $\mathcal{A'}$ cannot defend itself against all attacking arguments.
\end{itemize}
\end{definition}

The explanation presented in Figure \ref{fig:three-formats-explanations-abstract}(\subref{fig:argumentative-explanation}) with the edges representing support satisfies acceptability (as 
there are
no attacks).
The left-most and middle explanations in Figure~\ref{fig:dialectical-faithfulness-example} are acceptable as there is no argument attacking $a_4$, the only argument with conclusion $\hat{y}$.
Instead, the right-most explanation in Figure~\ref{fig:dialectical-faithfulness-example} is not acceptable, as there is no argument attacking $a_4$ in the explanation.
Note that the definition of acceptable could be extended to allow for chains of support to provide a defense, inspired by \citet{cayrol2005acceptability}.

\section{Evaluation Metrics}
We propose some metrics for evaluating empirically rationalizing explanations for NLP models, drawn from properties introduced in  \S~\ref{sec:properties}.\footnote{Implementations of the evaluation metrics discussed in this section can be found here: \url{https://github.com/neemakot/Evaluating-Explanations}}
We focus only on sample properties for lack of space, but metrics for other properties are also possible.

\subsection{Free-Form Evaluation}

We devise a metric for free-form explanation, $\textsc{Coh}$, relating to the property of coherence defined in \S~\ref{ssec:properties-unstructured}. For a free-form explanation  $\mathbf{E}=\mathcal{P}$ such that $\mathcal{P} \!\!\freeX m(\mathbf{X})\rightarrow \! \hat{y}$,  let 
$N = \vert\mathcal{P}\vert$ and $N'$ be the number of subsets of $\mathcal{P}$. Then, violation of
 coherence can be measured by $\textsc{Coh}(\mathbf{E})$ as shown in Eq.~\ref{eq:coherence}, where ${contr}(x,x')=1$ if $x$ is in contradiction with $x'$, and ${contr}(x,x')=0$ otherwise.

\begin{equation}
\footnotesize
\textsc{Coh}{(\mathbf{E})}\!= \!  \frac{1}{N'}{\!\sum_{{
\mathcal{P}' \subseteq \mathcal{P}}}{
({\neg contr}(
\mathcal{P}', \hat{y})
\cdot {\neg contr}(
\mathcal{P}', \mathbf{X}))}}
\label{eq:coherence}
\end{equation}

\subsection{Deductive Evaluation}
\label{sec:evaluating-structured}
We define the following metrics for deductive explanations: \textit{weak relevance}, \textit{strong relevance}, and \textit{redundancy}. We employ the properties presented in \S~\ref{sec:properties-structured} for deductive explanations to devise these metrics. We start with the metric $\textsc{Rel}_{\text{WEAK}}$ which is based on the weak relevance property. Let $\mathbf{E}=\langle \mathcal{P}, \mathcal{R}\rangle$ be a deductive explanation such that $\langle \mathcal{P}, \mathcal{R}\rangle \!\!\deductiveX\!\! m(\mathbf{X}) \!\!\rightarrow \! \hat{y}$, with $N = \vert\mathcal{P}\vert$. Then
satisfaction of weak relevance can be measured by:

\begin{equation}
\footnotesize
\textsc{Rel}_{\text{WEAK}}
(\mathbf{E}) = \frac{1}{N}\sum_{p_i \in \mathcal{P}}{path(p_i, \hat{y})}   
\end{equation}

where
$path(p_i, \hat{y})$ holds if there exists a path in $\mathcal{R}$ (seen as a graph) which connects $p_i$ to $\hat{y}$.
Note that $\textsc{Rel}_\text{WEAK}=0$ ($\textsc{Rel}_\text{WEAK}=1$) means that none (all, respectively) of the propositions in the explanation are relevant to the prediction.

For the related metric of strong relevance, we specify that there must be a relation, i.e. a direct connection between each proposition in the explanation and the prediction $\hat{y}$. This metric is defined as follows:

\begin{equation}
\footnotesize
\textsc{Rel}_{\text{STRONG}}
(\mathbf{E}) = \frac{1}{N}\sum_{p_i \in \mathcal{P}}{is\_relation(p_i, \hat{y})}   
\end{equation}

If all propositions are directly connected to $\hat{y}$, the value for this metric is $1$, whereas if none of the propositions is directly connected to $\hat{y}$, then $\textsc{Rel}_{\text{STRONG}}=0$.

The final metric that we provide for the evaluation of deductive explanations computes a score for the non-redundancy of an explanation. We define a non-redundant explanation in \S~\ref{sec:properties-structured} as one for which all propositions have relevance to the explanation in the context of the prediction. That is to say, if one of the propositions is omitted, the explanation would no longer be a sufficient justification for the model's prediction. The computation for the non-redundancy-derived metric is thus:
\begin{equation}
\footnotesize
\textsc{Red}(\mathbf{E}) = 1 - \frac{1}{N}\sum_{p_i \in \mathcal{P}}{(p_i \in \mathbf{X})  \cdot (p_i \in \mathcal{G_R})  }
\end{equation}

The non-redundancy metric checks that two conditions have been met. First, we must ensure that each proposition in the explanation is derived from the inputs, and second, it must be the case that there is some relation that connects each proposition to all others (either when the direction of edges is considered or not), i.e. if viewing the deductive explanation as a graph it should consist of a single connected component $\mathcal{G_R}$. The best possible score for redundancy is zero, indicating that the explanation contains no redundant components. A score greater than zero indicates at least one redundant proposition, if not more, exists in the explanation. 

\subsection{Argumentative Evaluation}
\label{sec:evaluating:argumentative}
We define metrics corresponding to two argumentative explanation properties: acceptability and dialectical non-circularity (see \S~\ref{sec:properties-argumentative-explanation}). For simplicity, we focus on argumentative explanations of a restricted kind, namely corresponding to sets of trees of depth two at most (where the root is of depth zero). Let $\mathbf{E}=\langle\mathcal{A}, \mathcal{R_{\text{Sup}}}, \mathcal{R_{\text{Att}}}\rangle$ be an argumentative explanation such that  $\langle\mathcal{A}, \mathcal{R_{\text{Sup}}}, \mathcal{R_{\text{Att}}}\rangle \argumentativeX m(\mathbf{X}) \rightarrow \hat{y}$, and let $N = \vert\mathcal{A}\vert $.
Then, the satisfaction of acceptability can be measured by 


\begin{equation}
\footnotesize
\textsc{Acc}(\mathbf{E})  = \frac{1}{N}{\sum_{{a_i = (\mathcal{P}, \hat{y}) \in \mathcal{A}}}}
(\frac{1}{\vert Atts(a_i)\vert }
     \sum_{(a_j, a_i) \in \mathcal{R}_{\text{Att}}} \delta(a_j))
\end{equation}

where $Atts(a_i)=\{a_j \mid (a_j, a_i) \in \mathcal{R}_{\text{Att}}$ and $\delta({a_j}))=1$ if
there exists 
$(a_k, a_j) \in \mathcal{R}_{\text{Att}}$, and  $\delta({a_j})=0$ otherwise. 
If $\hat{y}$ is predicted with top or high confidence we expect $\textsc{Acc}(\mathbf{E})=1$ for the explanation to be acceptable.
Instead, if $\hat{y}$ is predicted with bottom or low confidence, we expect $\textsc{Acc}(\mathbf{E})\neq 1$.

The second metric which we devise for evaluating argumentative explanations is related to the property of dialectical non-circularity. We define this metric as follows:

\begin{equation}
\footnotesize
    \textsc{Cir}(\mathbf{E})  = \frac{1}{N}{\sum_{ {a}\in \mathcal{A}} \frac{1}{M}{\sum_{\mathcal{A'} \in \mathcal{R_{\text{Sup}}},\mathcal{R_{\text{Att}}}}}  head(\mathcal{A'}, a) \cdot tail(\mathcal{A'}, a)}
\end{equation}

Here, we check for each argument in the explanation if there exists an attack or support relation in the explanation such that the head and tail arguments in the relation are the same arguments, i.e. the argumentation framework is circular. For the circularity measure, a favorable explanation would have a low score, i.e. fewer circular arguments.

\section{Conclusion}

We have identified and defined three rationale-derived explanation classes, drawing on illustrations from the automated fact-checking task in NLP. We also offered several desirable properties, both generic and structure-specific, for these explanations. Finally, we provided some quantitative measures for explanation evaluation. 

We understand that devising a framework that assumes that explanations modeled in the spirit of human reasoning will have some limitations, e.g. sociocultural differences in a population (one example being generational differences) may mean that an explanation that can be well understood by one population may not be as well received by another. In understanding this, our framework is modular and customizable, meaning it is flexible and can accommodate culturally and linguistically dependent preferences for explanations.

We would be interested to see the application of our metrics across a range of NLP tasks. 
Furthermore, there is also scope for expanding these metrics, either to account for further properties or to account for explanation structure at a finer level of granularity. Overall, we believe this work will help guide further research in explainable NLP and explanation evaluation.

\section{Acknowledgments}
We thank the anonymous reviewers for 
useful comments.
F. Toni was partially funded by the European Research Council (ERC) under
the European Union’s Horizon 2020 research and innovation programme (grant
agreement No. 101020934),  
by 
J.P. Morgan and  the
Royal Academy of Engineering under the Research Chairs
and Senior Research Fellowships scheme,
by the 
Royal Society, UK (IEC\textbackslash R2\textbackslash\- 222045 - International Exchanges 2022
) and by the UKRI INDICATE project.

\section{Ethics Statement}

In this paper, we present a framework for the evaluation of rationalizing explanations in the context of fact verification. We do not present empirical results, and, for that reason, we do not believe that serious ethical considerations arise from this work. However, we believe that this work presents a significant contribution towards improved AI ethics because explanations, and in particular means for assessing the quality of varied explanations, allow for greater model transparency and accountability. 

\section{Bibliographical References}
\label{sec:reference}
\bibliography{lrec-coling2024-example,anthology}
\bibliographystyle{lrec-coling2024-natbib}

\newpage

\end{document}